\documentclass{mva_style}
\usepackage{graphicx}
\usepackage{color}
\usepackage{amstext}
\usepackage{pifont}
\usepackage[subrefformat=parens]{subcaption}
\newcommand{\backvec}[1]{\reflectbox{\ensuremath{\vec{\reflectbox{\ensuremath{#1}}}}}}
\usepackage{array}
\usepackage{arydshln}

\usepackage{comment}
\usepackage{subcaption} 

\finalcopy 

\begin{document}
\title{Bidirectional Action Sequence Learning for Long-term Action Anticipation with Large Language Models}

\author{
  Yuji Sato \\
  Panasonic Connect Co., Ltd.\\
  {\tt sato.yuji@jp.panasonic.com}\\
  \and
  Yasunori Ishii \\
  Panasonic Holdings Corporation\\
  {\tt ishii.yasunori@jp.panasonic.com}\\
  \and
  Takayoshi Yamashita \\
  Chubu University\\
  {\tt takayoshi@isc.chubu.ac.jp}\\
}

\maketitle

\section*{\centering Abstract}
\vspace{-2mm}
Video-based long-term action anticipation is crucial for early risk detection in areas such as automated driving and robotics. Conventional approaches extract features from past actions using encoders and predict future events with decoders, which limits performance due to their unidirectional nature. These methods struggle to capture semantically distinct sub-actions within a scene. The proposed method, BiAnt, addresses this limitation by combining forward prediction with backward prediction using a large language model. Experimental results on Ego4D demonstrate that BiAnt improves performance in terms of edit distance compared to baseline methods.

\vspace{-2mm}
\section{Introduction}
\vspace{-2.5mm}
Video-based long-term action anticipation is important because it detects potential risks early.  
It has applications in reducing collision risks in automated driving, achieving safe motion control in robotics, and enabling early anomaly detection.

Previous studies~\cite{grauman2022ego4d, das2022video, ishibashi2023technical, gong2022future, nawhal2022rethinking, zhang2024object} extract features of past actions using convolutional networks or transformer encoders, and generate future action sequences via action classification.  
Recent methods employ large language models (LLMs)~\cite{zhao2024antgpt, huang2023palmpredictingactionslanguage}, embedding past action labels into prompts to improve prediction.  
While LLMs enable high-precision prediction even with diverse action labels, existing methods rely only on past actions and lack continuity in long-term sequences.  
This limits their ability to handle cases involving semantically distinct sub-actions, such as a person washing a knife and then serving pasta in a dish in a kitchen scene~\cite{grauman2022ego4d}.
Thus, modeling scene context that captures co-occurring but distinct actions is essential.

The proposed method, BiAnt, learns scene context by leveraging both past and future actions.  
It combines forward prediction (past to future) and backward prediction (future to past) to better understand action continuity.  
BiAnt trains the LLM using bidirectional prediction by leveraging both forward and backward prediction tasks to accurately predict future action sequences.
Given segmented videos, the method encodes past actions using a visual encoder, and generates prompts for both forward and backward anticipation tasks.  
Experiments on Ego4D~\cite{grauman2022ego4d} show that BiAnt improves edit distance over baselines that use only forward prediction.
The contributions of this paper are as follows:(1) The study presents BiAnt, a new framework that combines forward prediction tasks with backward prediction tasks. (2) The framework uses a large language model to predict action sequences and to learn bidirectional context. (3) Experiments on the Ego4D dataset show that BiAnt beats baseline methods in edit distance.
%
\begin{figure*}[t]
\centering
\scalebox{0.85}{
  \begin{minipage}{\linewidth}
    \centering
    \includegraphics[width=0.8\linewidth]{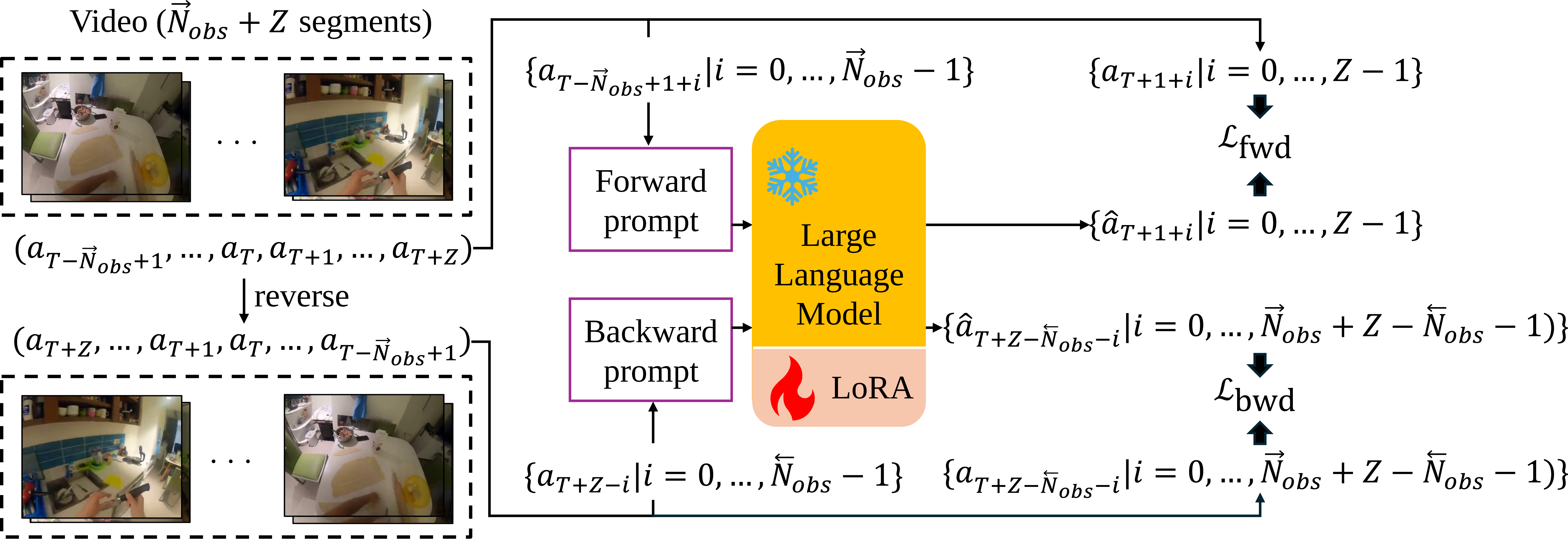} 
    \subcaption{\textbf{Training phase}}
    \label{fig:top}
  \end{minipage}
}
\scalebox{0.8}{
  \vspace{2mm} 
 \noindent\rule{\linewidth}{0.3pt}
  \vspace{2mm} 
}
\scalebox{0.85}{
  \begin{minipage}{\linewidth}
    \centering
    \includegraphics[width=0.8\linewidth]{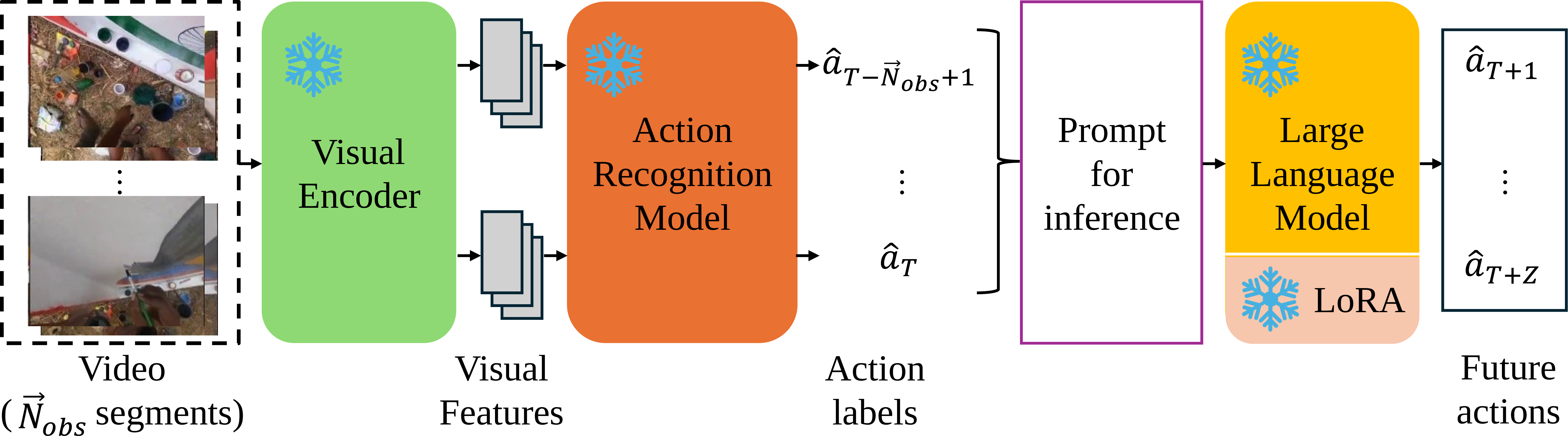} 
    \subcaption{\textbf{Inference phase}}
    \label{fig:bottom}
  \end{minipage}
  }
  \caption{
  \textbf{Overview of the BiAnt model architecture.} The model integrates a visual encoder, an action recognition model, and a large language model (LLM).
During training (a), the LLM learns from both forward and backward prompts.
During inference (b), only forward action labels are used to predict future actions. LoRA is applied for efficient LLM fine-tuning.
\color{black}}
\label{model_architecture}
\end{figure*}
\section{Related Work}
\vspace{-2.5mm}
\subsection{Action Anticipation}
\vspace{-2.5mm}
The objective of action anticipation is to predict future actions from an observed sequence.  
Tasks are divided into short-term, which predict the next action, and long-term, which predict subsequent action sequences.  
Long-term action anticipation is especially challenging due to the uncertainty and complexity of future actions.

Grauman \emph{et al.}~\cite{grauman2022ego4d} introduced the Ego4D dataset, comprising diverse first-person videos and proposed a baseline using convolutional neural networks (CNNs).  
Many studies adopted this dataset as a benchmark~\cite{mangalam2023egoschema, Kurita_2023_ICCV, Huang_2024_CVPR, Li_2024_CVPR, Zhu_2023_ICCV, zhao2023learning, grauman2024ego, bansal2022my, li2023ego, ohkawa2023assemblyhands, shan2020understanding}.  
The method~\cite{grauman2022ego4d} extracts features using SlowFast~\cite{feichtenhofer2019slowfast} or MViT~\cite{fan2021multiscale}, aggregates them, and classifies actions at each time step.
Transformer-based methods~\cite{gong2022future, nawhal2022rethinking} improve performance by modeling long-range dependencies.  
FUTR~\cite{gong2022future} uses an encoder-decoder structure to process past and future actions jointly.  
Anticipatr~\cite{nawhal2022rethinking} captures dependencies across segments with different actions using global and segment-level representations.
Recently, LLM-based methods~\cite{zhao2024antgpt, huang2023palmpredictingactionslanguage} have been proposed to leverage their generalization ability and contextual understanding.  
These methods generate labels for past actions using action recognition or captioning models, embed them into prompts, and use a LLM to predict future sequences.  
AntGPT~\cite{zhao2024antgpt} uses simple prompts with limited context, while Palm~\cite{huang2023palmpredictingactionslanguage} introduces more contextual information but increases computational complexity.
Unlike these LLM-based approaches that focus solely on forward prediction, our proposed method learns bidirectional action sequences using both forward and backward anticipation prompts.  
Although LLM-based methods remain underexplored in long-term action anticipation, our work introduces a new direction by effectively incorporating scene context through bidirectional modeling.
Unlike conventional language modeling where forward context dominates, action anticipation benefits from modeling reversed sequences to capture scene continuity.
We believe this work provides a foundation for advancing LLM-based research in action understanding.
\begin{figure*}[t]
\scalebox{1.0}{
\hspace{-3mm}
  \begin{tabular}{ccc}
    \captionsetup[subfigure]{font=small} 
    \begin{minipage}[t]{0.325\linewidth}
      \centering
      \includegraphics[width=\columnwidth]{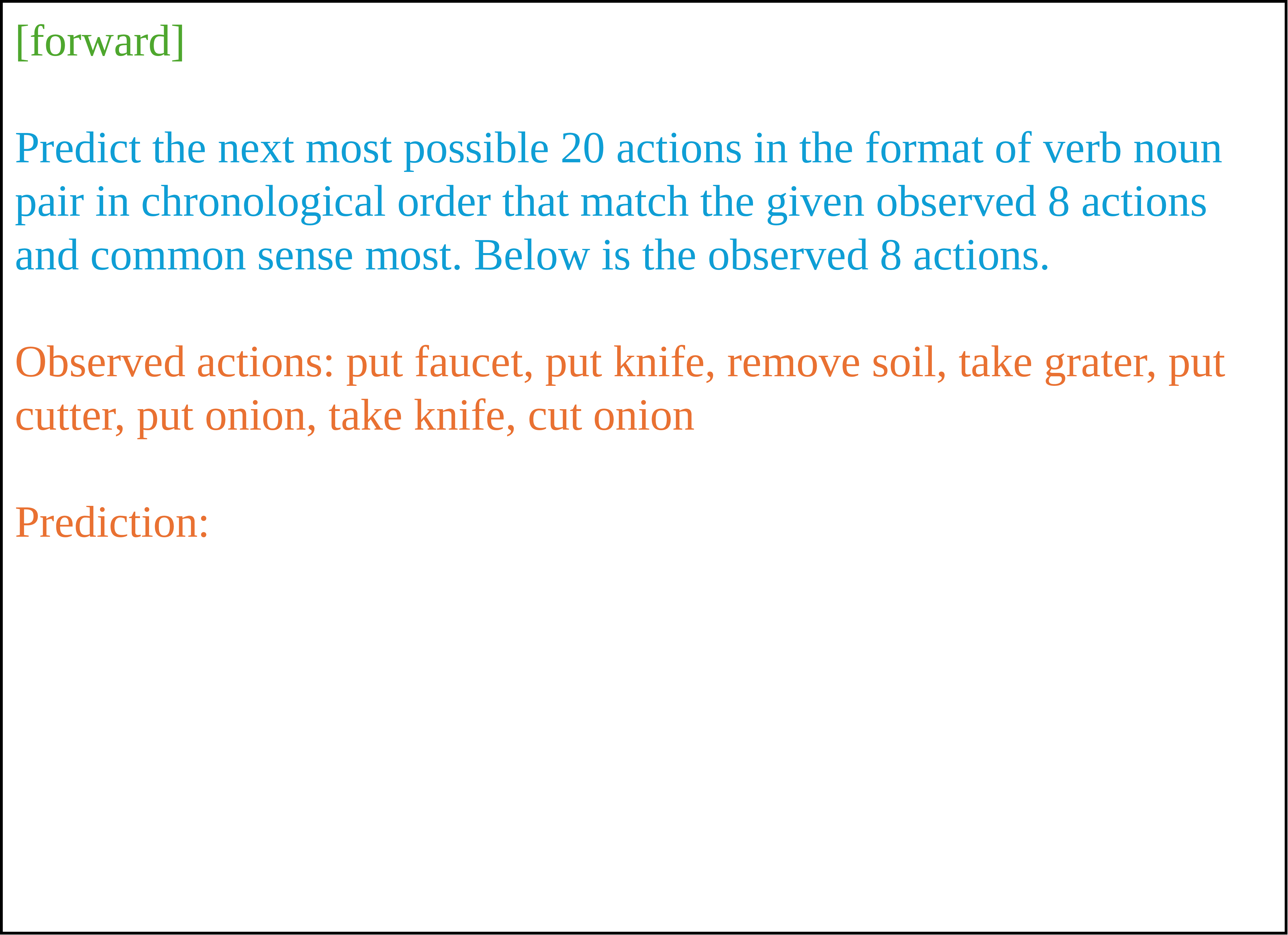}
      \subcaption{Forward action sequence learning}
      \label{fwd_prompt}
    \end{minipage}
%
    \begin{minipage}[t]{0.325\linewidth}
      \centering
      \includegraphics[width=\columnwidth]{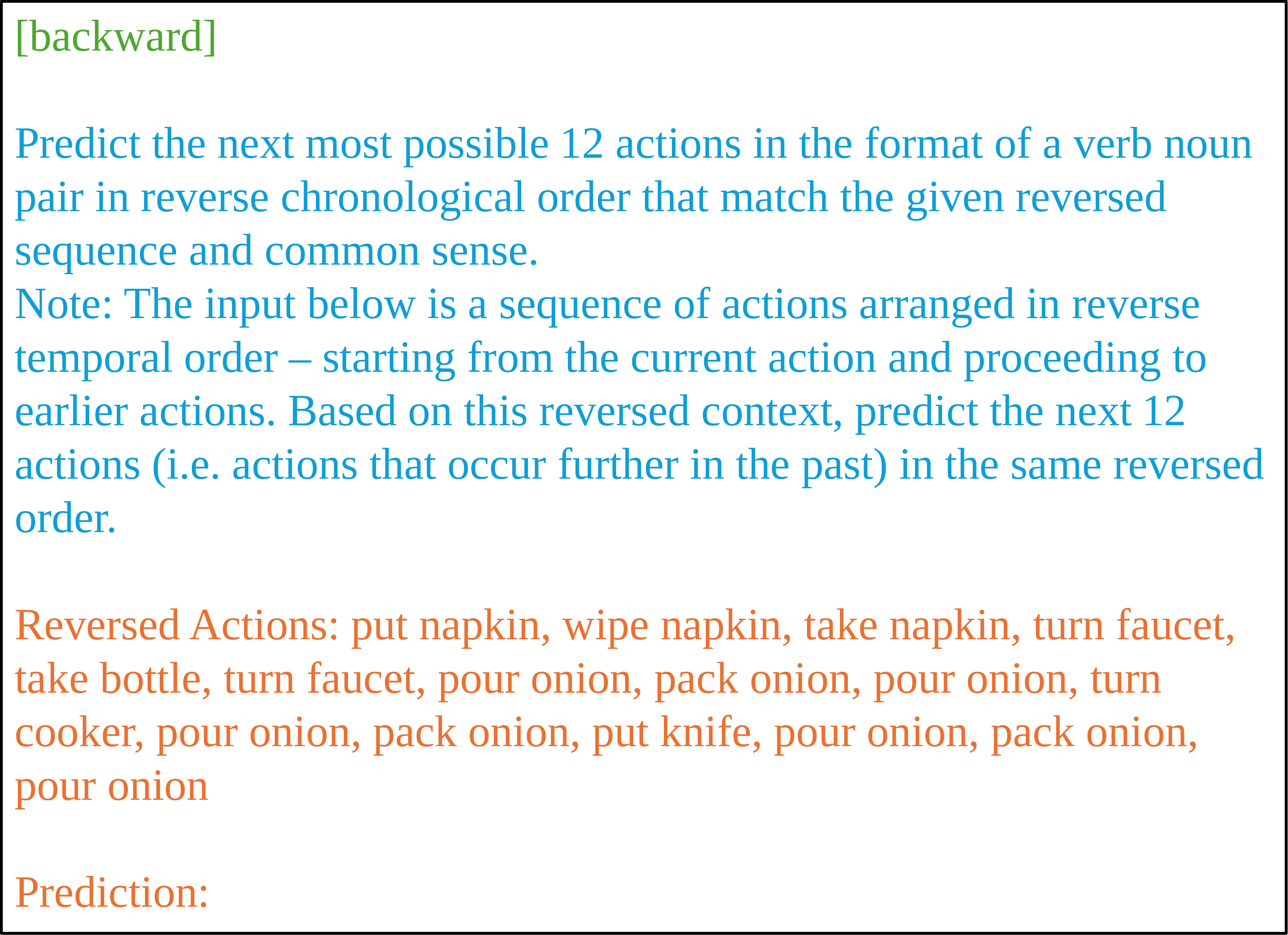}
     \subcaption{Backward action sequence learning}
      \label{bwd_prompt}
    \end{minipage}
%
    \begin{minipage}[t]{0.325\linewidth}
      \centering
      \includegraphics[width=\columnwidth]{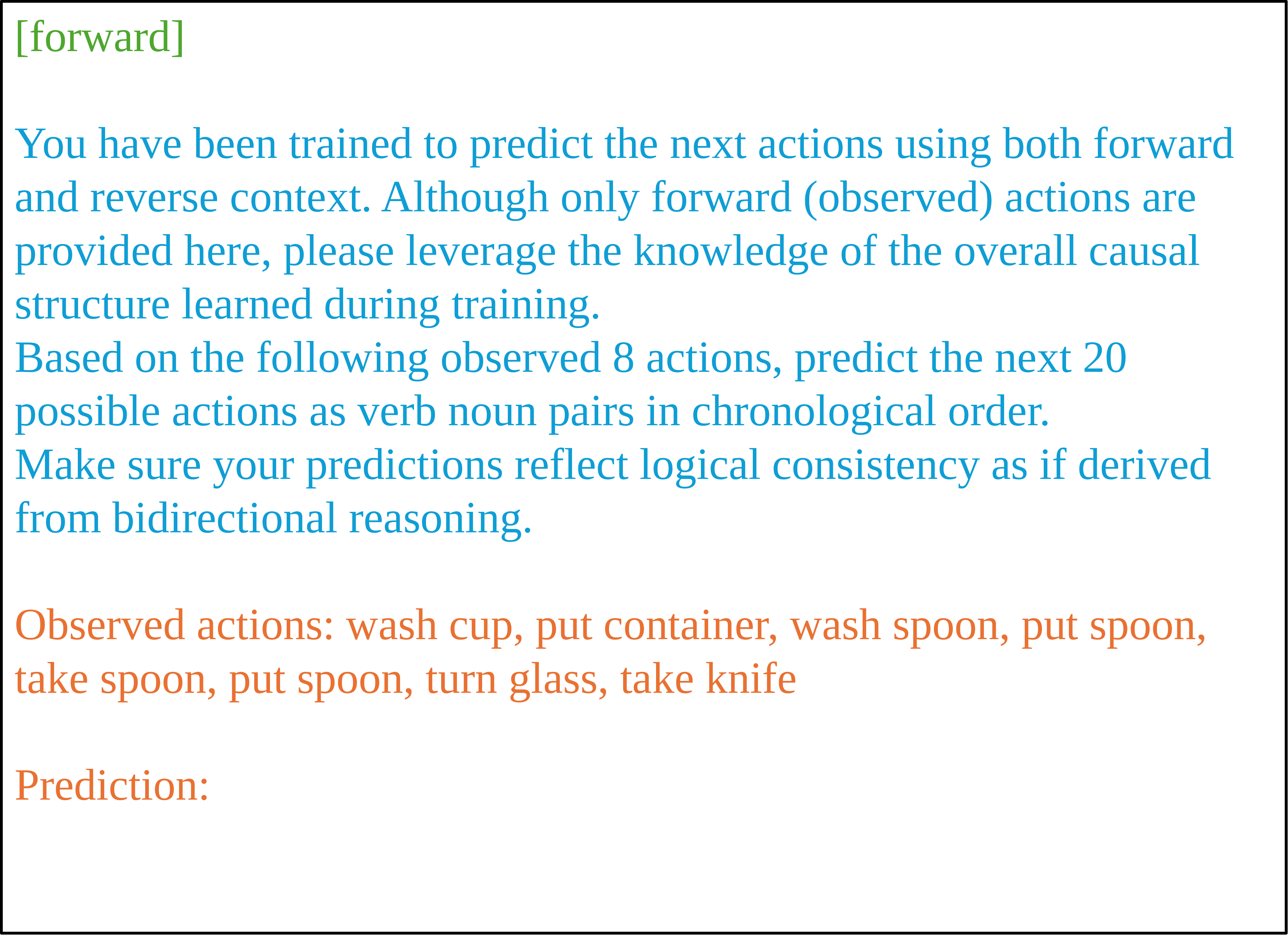}
     \subcaption{For inference}
      \label{inference_prompt}
    \end{minipage}

  \end{tabular}
    }
      \vspace{-2mm} 
  \caption{\textbf{Prompt design for action anticipation tasks.} 
  Examples of prompts used for (a) forward and (b) backward action sequence learning during training, and (c) forward-only inference.  
Each prompt includes a control token specifying the task type, an instruction guiding the LLM’s behavior, and a sequence of observed actions as input.  
  \color{black}
  }
  \label{prompt_design}
\end{figure*}
\vspace{-3.5mm}
\subsection{Bidirectional Learning}
\vspace{-3.5mm}
Bidirectional learning has been widely explored in natural language processing (NLP)~\cite{devlin2019bert, song2019mass, sanh2019distilbert, liu2019roberta, lanalbert, clarkelectra, he2020deberta}.  
BERT~\cite{devlin2019bert} introduced Masked Language Modeling (MLM), which captures both left and right contexts of each token to build deep bidirectional representations. 
Nguyen \emph{et al.}~\cite{nguyen2023meet} extended this idea using forward and backward decoders with a regularization term to align their predictions, improving infilling tasks via mutual learning.
Inspired by these approaches, we apply bidirectional learning to long-term action anticipation.  
Our method simultaneously learns forward and backward action sequence predictions, capturing dependencies between past and future actions.
\vspace{-2mm}
\section{BiAnt : Bidirectional Action Anticipation}
\vspace{-3mm}
\subsection{Bidirectional Training for Robust Forward-Only Action Sequence Generation}  
\vspace{-3mm}
Autoregressive generation with large language models (LLMs) predicts each next action based on previously generated actions.  
However, in sequences with sudden context shifts, even small errors can lead to cascading failures, making recovery difficult.  
To address this, we propose a bidirectional training method that jointly minimizes prediction errors in both forward and backward directions.  
By aligning forward and backward predictions during training, the model enforces consistency between both directions, suppressing incorrect predictions and reducing error cascades.

Importantly, although only forward prediction is used during inference, incorporating backward learning during training improves the model’s robustness and consistency.  
This is especially effective in dynamic video scenes with rapid action changes.  
Our method achieves this by modifying task descriptions in prompts, making it a simple yet effective solution for long-term action anticipation.
%
%
\vspace{-2.5mm}
\subsection{Model Architecture}
\vspace{-2.5mm}
Figure~\ref{model_architecture} shows the overall architecture of the proposed method, consisting of three modules: a visual encoder, an action recognition model, and a LLM.  
The visual encoder extracts features from input video segments, which the action recognition model uses to identify action labels.  
These modules are frozen during training.  
The LLM receives prompts embedded with action label sequences and performs both forward and backward action sequence prediction.  
Only the LLM is updated, using prediction errors from both directions.

In training (Figure~\ref{fig:top}), a video is divided into $N$ segments, each labeled with an action $a = (n, v)$, where $n$ and $v$ denote noun and verb.  
A stopping time $T$ defines the boundary between the observation interval of $\vec{N}_{obs}$ segments $(a^{T-\vec{N}_{obs}+1}, ..., a^T)$ and the future interval of $Z$ segments $(a^{T+1}, ..., a^{T+Z})$.

In the forward anticipation task, the observation sequence is embedded into a prompt, and the LLM generates the future action sequence.  
In the backward task, a reversed sequence of length $\vec{N}_{obs}+Z$ is constructed:
\[
(a^{T+Z}, ..., a^{T+1}, a^T, ..., a^{T-\vec{N}_{obs}+1})
\]
The first $\backvec{N}_{obs}$ actions form the reversed observation interval, and the remaining $\backvec{Z} = \vec{N}_{obs} + Z - \backvec{N}_{obs}$ form the reversed future.  
This reversed observation is used as a prompt to predict the reversed future.  
Losses from both forward and backward outputs are computed against ground truth, and the LLM is updated to minimize them jointly.

During inference (Figure~\ref{fig:bottom}), given $\vec{N}_{obs}$ segments, the visual encoder extracts features, and the action recognition model assigns labels to the observation interval.  
These labels are embedded into a prompt, and the LLM generates the future action sequence using forward prediction only.
\vspace{-3mm}
\subsection{Bidirectional Action Sequence Learning}
\vspace{-3.5mm}
Conventional training for action prediction uses a prompt consisting of a task instruction and a past action label sequence to train the LLM for future action generation.  
Our method extends this setup to include a \textbf{backward anticipation task}, enabling the model to learn bidirectional dependencies between past and future actions.

As shown in Figure~\ref{fwd_prompt} and Figure~\ref{bwd_prompt}, each prompt begins with a control token that specifies the task type, forward or backward, allowing the LLM to learn task-specific behavior.  
The backward prompt additionally includes a brief instruction explaining the use of reversed sequences, guiding the model to generate appropriate predictions during training.
%
%

\noindent
\textbf{Loss Calculation:}  
In the \textbf{forward anticipation} task, the model minimizes the loss between the predicted future action sequence (based on the observation interval) and the ground-truth sequence:
\begin{equation}
  \label{fwd_loss}
  \mathcal{L}_{\text{fwd}} = \sum_{i=1}^{Z} \text{CrossEntropy} (\vec{y}_{i}, \vec{p}_{i}),
\end{equation}
where $Z$ is the prediction length, $\vec{y}_{i}$ is the ground-truth labels, and $\vec{p}_{i}$ is the predicted probability at position $i$.

Similarly, the \textbf{backward anticipation} task minimizes the loss between the reversed predicted sequence and the reversed ground-truth:
\begin{equation}
  \label{bwd_loss}
  \mathcal{L}_{\text{bwd}} = \sum_{i=1}^{\backvec{Z}} \text{CrossEntropy} (\backvec{y}_{i}, \backvec{p}_{i}),
\end{equation}
where $\backvec{Z}$ is the prediction length, $\backvec{y}_{i}$ and $\backvec{p}_{i}$ denote the labels and predicted probability in the reversed sequence.

The overall training loss is a weighted combination of both:
\begin{equation}
  \label{total_loss}
  \mathcal{L} = \alpha \mathcal{L}_{\text{fwd}} + \beta \mathcal{L}_{\text{bwd}},
\end{equation}
where $\alpha$ and $\beta$ are weighting coefficients for each task.
\vspace{-6mm}
\subsection{Inference}
\vspace{-3mm}
Figure~\ref{inference_prompt} shows the prompt used during inference.  
In this phase, the visual encoder and action recognition model extract the action sequence from the observation interval.  
This sequence is then embedded into a prompt, and the LLM generates the future action sequence.  
As in training, a control token at the beginning specifies the task as forward anticipation.
\color{red}
\color{black}
\vspace{-2.5mm}
\section{Experiments}
\vspace{-3.5mm}
\noindent
\textbf{Dataset and Evaluation Setting:}  
We use the Ego4D dataset~\cite{grauman2022ego4d}, a large-scale collection of first-person videos capturing diverse daily activities.  
Each video is divided into segments labeled with combinations of 117 verbs and 521 nouns, with each segment containing about 240 frames.

Evaluation follows the Ego4D protocol, using Edit Distance (ED) to assess prediction accuracy for nouns, verbs, and actions.  
From $K=5$ generated candidates, the sequence with the lowest ED is selected.  
We set the observation interval length to $\vec{N}_{obs} = 8$ and the future prediction length to $Z = 20$, following~\cite{grauman2022ego4d}.
\noindent
\textbf{Implementation Details:}  
We adopt AntGPT~\cite{zhao2024antgpt} as the baseline and use Llama2-7B as the LLM with parameter-efficient fine-tuning.  
Training is conducted with a batch size of 32, learning rate of 3e-4, for 3 epochs using 8-bit quantization.  
Loss weights are set to $\alpha = 1.0$ and $\beta = 1.0$.
For the reversed action sequence, $\backvec{N}_{obs}$ is set to 16, and $\backvec{Z}$ is set to 12.
\\
\noindent
\textbf{Quantitative Evaluation:}
Table~\ref{ego4d_v2_test_result} presents the results on the Ego4D v2 test set.  
Action label-embedded LLM-based methods (Palm~\cite{huang2023palmpredictingactionslanguage}, AntGPT~\cite{zhao2024antgpt}) outperform other approaches.
Although verbs and nouns are predicted, the key metric is action.
Our proposed method further improves performance, particularly on the action metric.  
This demonstrates that training with both forward and backward anticipation enhances the model’s ability to leverage bidirectional context, even when using only forward inputs during inference.
%
\begin{table}[t]
  \renewcommand{\arraystretch}{1.2}
  \caption{Quantitative evaluation on Ego4D v2 test set.}
  \begin{center}
  \vspace{-6mm}
\scalebox{0.75}{
    \begin{tabular}{c | c c c}
      \hline
      \hline
      \makebox[10mm]{Method} & \makebox[10mm]{verb $\downarrow$} &
      \makebox[10mm]{noun $\downarrow$} & \makebox[10mm]{action $\downarrow$}\\
      \hline
      SlowFast~\cite{feichtenhofer2019slowfast} & 0.7169 & 0.7359 & 0.9253 \\
      VideoLLM~\cite{chen2023videollm} & 0.7210 & 0.7250 & 0.9210 \\
      PsMsEgoAI~\cite{ishibashi2023technical} & 0.6838 & 0.6785 & 0.8933 \\
      Palm~\cite{huang2023palmpredictingactionslanguage} & 0.6956 & 0.6506 & 0.8856 \\
      AntGPT~\cite{zhao2024antgpt} & 0.6503 & 0.6498 & 0.8770 \\
      \hdashline
      BiAnt (ours) & \bf{0.6377} & \bf{0.6468} & \bf{0.8655} \\
      \hline
      \hline
    \end{tabular}
    }
    \label{ego4d_v2_test_result}
  \end{center}
\end{table}
\begin{table}
  \renewcommand{\arraystretch}{1.2}
  \caption{Ablation on task description token}
  \begin{center}
    \vspace{-6mm}
  \scalebox{0.75}{
    \begin{tabular}{c | c c c}
      \hline
      \hline
      \makebox[33mm]{Task description token} & 
      \makebox[10mm]{verb $\downarrow$} & \makebox[10mm]{noun $\downarrow$} & \makebox[10mm]{action $\downarrow$} \\
      \hline
      \ding{51} & 0.6841 & \bf{0.6378} & \bf{0.8741} \\
        & \bf{0.6747} & 0.6425 & 0.8866 \\
      \hline
      \hline
    \end{tabular}
    }
    \label{ablation_study}
  \end{center}
\end{table}
\vspace{-1mm}
\begin{figure}[t]
\centering
\scalebox{0.98}{
   \captionsetup[subfigure]{font=small} 
      \centering
\hspace{-4mm}
  \begin{tabular}{cc}
    \begin{minipage}[t]{1.0\linewidth}
      \includegraphics[width=\columnwidth]{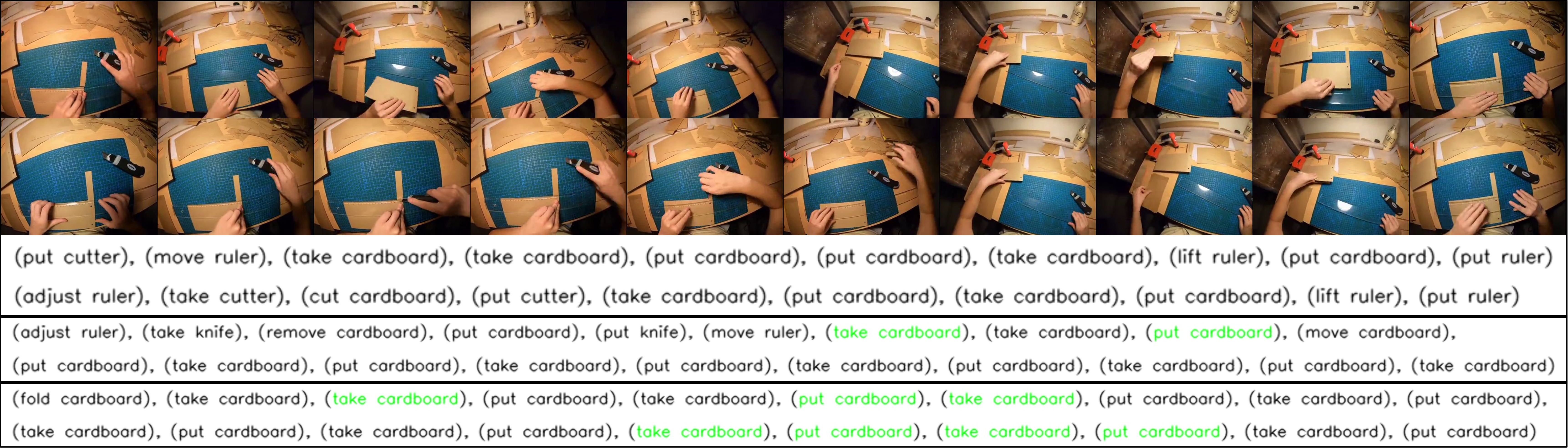}
\color{blue}
\vspace{-5mm}
      \subcaption{From top: the evaluation video, the GT action sequence, and the estimated action sequences from 
      AntGPT (ED = 0.7)   and Ours (ED = 0.6).}
\color{black}
      \label{qualitative_result_desk}
    \end{minipage}\\
    \begin{minipage}[t]{1.0\linewidth}
      \centering
  \includegraphics[width=\columnwidth]{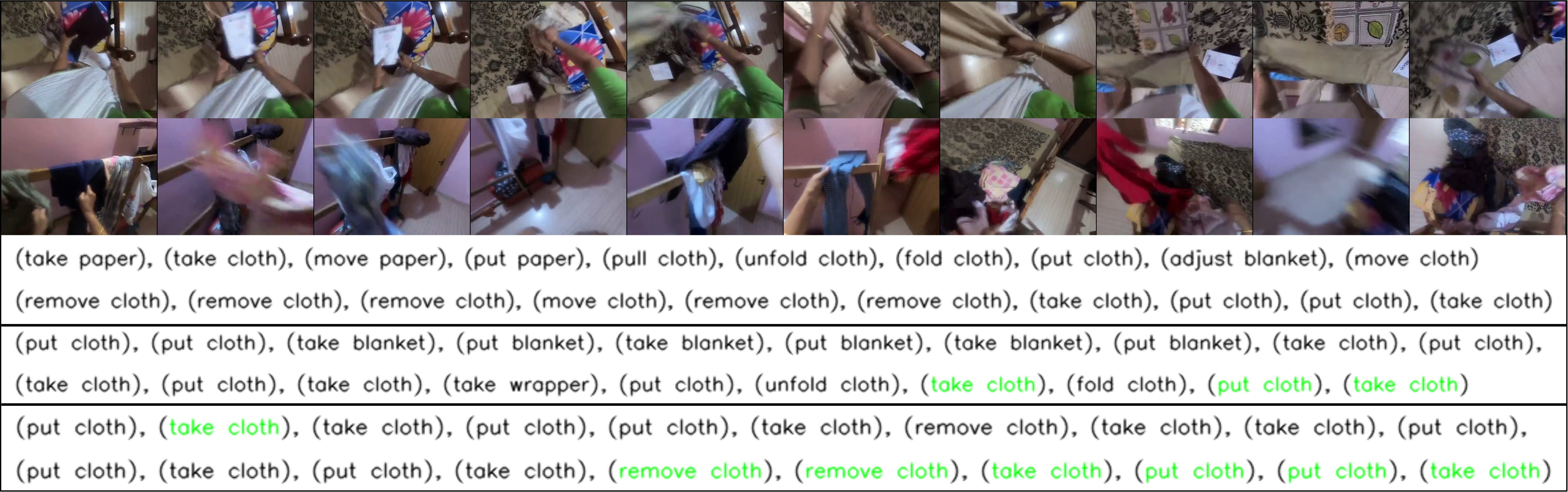}
\color{blue}
\vspace{-5mm}
      \subcaption{From top: the evaluation video, the GT action sequence, and the estimated action sequences from AntGPT (ED = 0.8) and Ours (ED = 0.65).}
\color{black}
      \label{qualitative_result_laundry}
    \end{minipage}
  \end{tabular}
  }
  \vspace{-3mm}
  \caption{
\textbf{Qualitative comparisons of predicted action sequences.}
Results for (a) a desk-working scene and (b) a laundry scene in which three actions are predominantly repeated.
Ground-truth (GT) sequences are shown at the top, followed by predictions from AntGPT and the proposed method (BiAnt).  
Correct predictions are highlighted in green. BiAnt achieves lower Edit Distances (0.6 vs. 0.7 in (a), 0.65 vs. 0.8 in (b)), indicating improved accuracy.
\color{black}}
  \label{qualitative_result}
\end{figure}
%
%
\vspace{-3mm}
\\
\noindent
\textbf{Ablation Study:}
We conduct an ablation study to evaluate the effect of the task description token, comparing results with and without it on the Ego4D validation subset (160 samples from the official split).
As shown in Table~\ref{ablation_study}, explicitly specifying the task type helps the model learn each direction effectively.  
Adding the task token improves the action ED score by 0.0125pt, demonstrating its role in guiding the model to better utilize bidirectional context during training.

\noindent
\textbf{Qualitative Analysis:}
Figures~\ref{qualitative_result_desk} and \ref{qualitative_result_laundry} present qualitative comparisons between AntGPT~\cite{zhao2024antgpt} and the proposed method. In Figure~\ref{qualitative_result_desk}, the scene depicts the repeated action of cutting cardboard on a desk.
AntGPT predicts actions within specific segments, whereas the proposed method successfully predicts the recurring sequence of actions, such as taking and putting a cardboard. In Figure~\ref{qualitative_result_laundry}, the scene depicts household chores that primarily involve action labels such as remove cloth, take cloth, and put cloth. AntGPT predicts individual actions, such as take or put cloth, but fails to capture the complete action flow. In contrast, the proposed method accurately predicts the entire sequence of actions—including later, more complex segments—thereby demonstrating its strength in modeling long-range action dependencies.

\vspace{-3mm}
\section{Conclusion}
\vspace{-3.5mm}
We proposed BiAnt, a bidirectional framework for long-term action anticipation in video.  
By combining forward and backward prediction, BiAnt effectively captures scene context from both past and future.  
Experiments on Ego4D confirmed improved performance, and ablation results highlighted the impact of task tokens.  
These findings show that bidirectional learning enhances the prediction of complex human actions.
While LLM-based action anticipation is still an emerging field, 
this study demonstrates that incorporating backward action sequences allows models to better capture video context beyond conventional forward modeling.  
Although there remains room for improving how context is learned from backward sequences, we believe our work offers a foundational idea that will inspire further research in long-term action anticipation.

\color{black}
{
    \small
    \bibliographystyle{plain}
    \bibliography{main}

\begin{thebibliography}{10}

\bibitem{bansal2022my}
Siddhant Bansal, Chetan Arora, and CV~Jawahar.
\newblock My view is the best view: Procedure learning from egocentric videos.
\newblock In {\em {Proceedings of the European Conference on Computer Vision (ECCV)}}, pages 657--675. Springer, 2022.

\bibitem{chen2023videollm}
Guo Chen, Yin-Dong Zheng, Jiahao Wang, Jilan Xu, Yifei Huang, Junting Pan, Yi~Wang, Yali Wang, Yu~Qiao, Tong Lu, and Limin Wang.
\newblock {VideoLLM}: Modeling video sequence with large language models.
\newblock {\em arXiv preprint arXiv:2305.13292}, 2023.

\bibitem{clarkelectra}
Kevin Clark, Minh-Thang Luong, Quoc~V Le, and Christopher~D Manning.
\newblock {ELECTRA}: Pre-training text encoders as discriminators rather than generators.
\newblock In {\em {Proceedings of the International Conference on Learning Representations (ICLR)}}, 2020.

\bibitem{das2022video}
Srijan Das and Michael~S. Ryoo.
\newblock Video + clip baseline for ego4d long-term action anticipation.
\newblock {\em arXiv preprint arXiv:2207.00579}, 2022.

\bibitem{devlin2019bert}
Jacob Devlin, Ming-Wei Chang, Kenton Lee, and Kristina Toutanova.
\newblock {BERT}: Pre-training of deep bidirectional transformers for language understanding.
\newblock In {\em {Proceedings of the North American Chapter of the Association for Computational Linguistics (NAACL)}}, pages 4171--4186, 2019.

\bibitem{fan2021multiscale}
Haoqi Fan, Bo~Xiong, Karttikeya Mangalam, Yanghao Li, Zhicheng Yan, Jitendra Malik, and Christoph Feichtenhofer.
\newblock Multiscale vision transformers.
\newblock In {\em {Proceedings of the IEEE International Conference on Computer Vision (ICCV)}}, pages 6824--6835, 2021.

\bibitem{feichtenhofer2019slowfast}
Christoph Feichtenhofer, Haoqi Fan, Jitendra Malik, and Kaiming He.
\newblock Slowfast networks for video recognition.
\newblock In {\em {Proceedings of the IEEE International Conference on Computer Vision (ICCV)}}, pages 6202--6211, 2019.

\bibitem{gong2022future}
Dayoung Gong, Joonseok Lee, Manjin Kim, Seong~Jong Ha, and Minsu Cho.
\newblock Future transformer for long-term action anticipation.
\newblock In {\em {Proceedings of the IEEE Conference on Computer Vision and Pattern Recognition (CVPR)}}, pages 3052--3061, 2022.

\bibitem{grauman2022ego4d}
Kristen Grauman, Andrew Westbury, Eugene Byrne, Zachary Chavis, Antonino Furnari, Rohit Girdhar, Jackson Hamburger, Hao Jiang, Miao Liu, Xingyu Liu, et~al.
\newblock {Ego4D}: Around the world in 3,000 hours of egocentric video.
\newblock In {\em {Proceedings of the IEEE Conference on Computer Vision and Pattern Recognition (CVPR)}}, pages 18995--19012, 2022.

\bibitem{grauman2024ego}
Kristen Grauman, Andrew Westbury, Lorenzo Torresani, Kris Kitani, Jitendra Malik, Triantafyllos Afouras, Kumar Ashutosh, Vijay Baiyya, Siddhant Bansal, Bikram Boote, et~al.
\newblock {Ego-Exo4D}: Understanding skilled human activity from first-and third-person perspectives.
\newblock In {\em {Proceedings of the IEEE Conference on Computer Vision and Pattern Recognition (CVPR)}}, pages 19383--19400, 2024.

\bibitem{he2020deberta}
Pengcheng He, Xiaodong Liu, Jianfeng Gao, and Weizhu Chen.
\newblock {DeBERTa}: Decoding-enhanced bert with disentangled attention.
\newblock In {\em {Proceedings of the International Conference on Learning Representations (ICLR)}}, 2020.

\bibitem{huang2023palmpredictingactionslanguage}
Daoji Huang, Otmar Hilliges, Luc~Van Gool, and Xi~Wang.
\newblock {PALM}: Predicting actions through language models @ ego4d long-term action anticipation challenge 2023.
\newblock {\em arXiv preprint arXiv:2306.16545}, 2023.

\bibitem{Huang_2024_CVPR}
Yifei Huang, Guo Chen, Jilan Xu, Mingfang Zhang, Lijin Yang, Baoqi Pei, Hongjie Zhang, Lu~Dong, Yali Wang, Limin Wang, and Yu~Qiao.
\newblock {EgoExoLearn}: A dataset for bridging asynchronous ego- and exo-centric view of procedural activities in real world.
\newblock In {\em {Proceedings of the IEEE Conference on Computer Vision and Pattern Recognition (CVPR)}}, pages 22072--22086, June 2024.

\bibitem{ishibashi2023technical}
Tatsuya Ishibashi, Kosuke Ono, Noriyuki Kugo, and Yuji Sato.
\newblock Technical report for ego4d long term action anticipation challenge 2023.
\newblock {\em arXiv preprint arXiv:2307.01467}, 2023.

\bibitem{Kurita_2023_ICCV}
Shuhei Kurita, Naoki Katsura, and Eri Onami.
\newblock {RefEgo}: Referring expression comprehension dataset from first-person perception of ego4d.
\newblock In {\em {Proceedings of the IEEE International Conference on Computer Vision (ICCV)}}, pages 15214--15224, October 2023.

\bibitem{lanalbert}
Zhenzhong Lan, Mingda Chen, Sebastian Goodman, Kevin Gimpel, Piyush Sharma, and Radu Soricut.
\newblock {ALBERT}: A lite bert for self-supervised learning of language representations.
\newblock In {\em {Proceedings of the International Conference on Learning Representations (ICLR)}}, 2020.

\bibitem{Li_2024_CVPR}
Gen Li, Kaifeng Zhao, Siwei Zhang, Xiaozhong Lyu, Mihai Dusmanu, Yan Zhang, Marc Pollefeys, and Siyu Tang.
\newblock {EgoGen}: An egocentric synthetic data generator.
\newblock In {\em {Proceedings of the IEEE Conference on Computer Vision and Pattern Recognition (CVPR)}}, pages 14497--14509, June 2024.

\bibitem{li2023ego}
Jiaman Li, Karen Liu, and Jiajun Wu.
\newblock Ego-body pose estimation via ego-head pose estimation.
\newblock In {\em {Proceedings of the IEEE Conference on Computer Vision and Pattern Recognition (CVPR)}}, pages 17142--17151, 2023.

\bibitem{liu2019roberta}
Yinhan Liu, Myle Ott, Naman Goyal, Jingfei Du, Mandar Joshi, Danqi Chen, Omer Levy, Mike Lewis, Luke Zettlemoyer, and Veselin Stoyanov.
\newblock {RoBERTa}: A robustly optimized bert pretraining approach.
\newblock {\em arXiv preprint arXiv:1907.11692}, 2019.

\bibitem{mangalam2023egoschema}
Karttikeya Mangalam, Raiymbek Akshulakov, and Jitendra Malik.
\newblock {EgoSchema}: A diagnostic benchmark for very long-form video language understanding.
\newblock {\em {Proceedings of the Advances in Neural Information Processing Systems (NeurIPS)}}, 36:46212--46244, 2023.

\bibitem{nawhal2022rethinking}
Megha Nawhal, Akash~Abdu Jyothi, and Greg Mori.
\newblock Rethinking learning approaches for long-term action anticipation.
\newblock In {\em {Proceedings of the European Conference on Computer Vision (ECCV)}}, pages 558--576. Springer, 2022.

\bibitem{nguyen2023meet}
Anh Nguyen, Nikos Karampatziakis, and Weizhu Chen.
\newblock Meet in the middle: A new pre-training paradigm.
\newblock {\em {Proceedings of the Advances in Neural Information Processing Systems (NeurIPS)}}, 36:5079--5091, 2023.

\bibitem{ohkawa2023assemblyhands}
Takehiko Ohkawa, Kun He, Fadime Sener, Tomas Hodan, Luan Tran, and Cem Keskin.
\newblock {AssemblyHands}: Towards egocentric activity understanding via 3d hand pose estimation.
\newblock In {\em {Proceedings of the IEEE Conference on Computer Vision and Pattern Recognition (CVPR)}}, pages 12999--13008, 2023.

\bibitem{sanh2019distilbert}
Victor Sanh, Lysandre Debut, Julien Chaumond, and Thomas Wolf.
\newblock {DistilBERT}, a distilled version of bert: smaller, faster, cheaper and lighter.
\newblock {\em arXiv preprint arXiv:1910.01108}, 2019.

\bibitem{shan2020understanding}
Dandan Shan, Jiaqi Geng, Michelle Shu, and David~F Fouhey.
\newblock Understanding human hands in contact at internet scale.
\newblock In {\em {Proceedings of the IEEE Conference on Computer Vision and Pattern Recognition (CVPR)}}, pages 9869--9878, 2020.

\bibitem{song2019mass}
Kaitao Song, Xu~Tan, Tao Qin, Jianfeng Lu, and Tie-Yan Liu.
\newblock {MASS}: Masked sequence to sequence pre-training for language generation.
\newblock In {\em {Proceedings of the International Conference on Machine Learning (ICML)}}, pages 5926--5936. PMLR, 2019.

\bibitem{zhang2024object}
Ce~Zhang, Changcheng Fu, Shijie Wang, Nakul Agarwal, Kwonjoon Lee, Chiho Choi, and Chen Sun.
\newblock Object-centric video representation for long-term action anticipation.
\newblock In {\em {Proceedings of the IEEE Winter Conference on Applications of Computer Vision (WACV)}}, 2024.

\bibitem{zhao2024antgpt}
Qi~Zhao, Shijie Wang, Ce~Zhang, Changcheng Fu, Minh~Quan Do, Nakul Agarwal, Kwonjoon Lee, and Chen Sun.
\newblock {AntGPT}: Can large language models help long-term action anticipation from videos?
\newblock In {\em {Proceedings of the International Conference on Learning Representations (ICLR)}}, 2024.

\bibitem{zhao2023learning}
Yue Zhao, Ishan Misra, Philipp Kr{\"a}henb{\"u}hl, and Rohit Girdhar.
\newblock Learning video representations from large language models.
\newblock In {\em {Proceedings of the IEEE Conference on Computer Vision and Pattern Recognition (CVPR)}}, pages 6586--6597, 2023.

\bibitem{Zhu_2023_ICCV}
Chenchen Zhu, Fanyi Xiao, Andres Alvarado, Yasmine Babaei, Jiabo Hu, Hichem El-Mohri, Sean Culatana, Roshan Sumbaly, and Zhicheng Yan.
\newblock {EgoObjects}: A large-scale egocentric dataset for fine-grained object understanding.
\newblock In {\em {Proceedings of the IEEE International Conference on Computer Vision (ICCV)}}, pages 20110--20120, October 2023.

\end{thebibliography}


\begin{thebibliography}{1}

\bibitem{grauman2022ego4d}
Kristen Grauman, Andrew Westbury, Eugene Byrne, Zachary Chavis, Antonino Furnari, Rohit Girdhar, Jackson Hamburger, Hao Jiang, Miao Liu, Xingyu Liu, et~al.
\newblock {Ego4D}: Around the world in 3,000 hours of egocentric video.
\newblock In {\em {Proceedings of the IEEE Conference on Computer Vision and Pattern Recognition (CVPR)}}, pages 18995--19012, 2022.

\end{thebibliography}
}

\end{document}


\title{Bidirectional Action Sequence Learning for Long-term Action Anticipation with Large Language Models \\ (Supplementary Material)}

\maketitle

\section{Ablation Study}
This section presents an ablation study on the model design in Ego4D~\cite{grauman2022ego4d}.
The study evaluates the effects of different observation interval lengths and loss weights on the performance of the backward anticipation task and investigates the impact of various task description tokens placed at the beginning of the input prompt.
In this experiment, the edit distance is used for comparative evaluation, and three metrics—verb, noun, and action—are reported.

\subsection{Length of the observation interval for the backward anticipation task}
The observation interval in the forward anticipation task must align with other methods; however, the proposed backward anticipation task can be determined independently of the observation interval in the forward anticipation task. Therefore, a comparison of the impact of the observation interval on performance is conducted.
To investigate the influence of the observation interval length on the backward anticipation task, four different lengths—4, 8, 16, and 24—were evaluated. Length 8 is the default observation interval in the Ego4D dataset. Table \ref{ablation_study_on_observation_interval} presents the accuracy as the observation interval length varies, and the highest accuracy for the action metric is achieved when the observation interval is set to 16.
When only 4 or 8 segments are observed, the model lacks enough temporal information for reliable predictions. Training with 24 segments, however, creates a large gap compared to the 8 segments used during inference. In contrast, using 16 segments provides sufficient temporal information while closely matching the inference conditions, resulting in the highest performance.

\subsection{Loss weights}
%
The proposed method simultaneously learns the forward anticipation task and the backward anticipation task; however, inference occurs through the forward anticipation task.
The investigation examines the impact of using the backward anticipation task in learning by varying the loss weight, $\beta$.
To investigate the influence of the loss weights $\alpha$ and $\beta$, three different combinations were evaluated: $\alpha = 1.0$, $\beta = 0.5$; $\alpha = 1.0$, $\beta = 0.75$; and $\alpha = 1.0$, $\beta = 1.0$. Here, $\alpha$ and $\beta$ are the weights for the loss functions of the forward and backward anticipation tasks, respectively. Table \ref{ablation_study} presents the accuracy as the loss weights $\alpha$ and $\beta$ are varied. The highest accuracy across all metrics is observed when $\alpha$ is set to 1.0 and $\beta$ is set to 1.0.

\begin{table}[t]
  \renewcommand{\arraystretch}{1.2}
  \caption{
  Ablation on the observation interval length for the backward anticipation task. The best accuracy is highlighted in bold. A length of 16 achieves the highest accuracy for the action metric.
  }
  \begin{center}
  \scalebox{0.8}{
    \begin{tabular}{c | c c c}
      \hline
      \hline
      \makebox[10mm]{Length} &
      \makebox[10mm]{verb $\downarrow$} & \makebox[10mm]{noun $\downarrow$} & \makebox[10mm]{action $\downarrow$} \\
      \hline
      4 & 0.6822 & \bf{0.6281} & 0.8822 \\
      8 & \bf{0.6772} & 0.6622 & 0.8872 \\
      16 & 0.6841 & 0.6378 & \bf{0.8741} \\
      24 & 0.6909 & 0.6381 & 0.8847 \\
      \hline
      \hline
    \end{tabular}
    }
    \label{ablation_study_on_observation_interval}
  \end{center}
\end{table}
\begin{table}[t]
  \renewcommand{\arraystretch}{1.2}
  \caption{
  Ablation on the different values of loss weights $\alpha$ and $\beta$. The best accuracy is highlighted in bold. The optimal accuracy is achieved when $\alpha$ is set to 1.0 and $\beta$ is set to 1.0.
  }
  \begin{center}
  \scalebox{0.8}{
    \begin{tabular}{c c | c c c}
      \hline
      \hline
      \makebox[10mm]{$\alpha$} & \makebox[10mm]{$\beta$} &
      \makebox[10mm]{verb $\downarrow$} & \makebox[10mm]{noun $\downarrow$} & \makebox[10mm]{action $\downarrow$} \\
      \hline
      1.0 & 0.5 & 0.6856 & 0.6512 & 0.8828 \\
      1.0 & 0.75 & 0.6928 & 0.6466 & 0.8853 \\
      1.0 & 1.0 & \bf{0.6841} & \bf{0.6378} & \bf{0.8741} \\
      \hline
      \hline
    \end{tabular}
    }
    \label{ablation_study}
  \end{center}
\end{table}
%
\begin{figure*}[t]
  \scalebox{1.0}{
  \hspace{-3mm}
    \begin{tabular}{ccc}
      \captionsetup[subfigure]{font=small} 
      \begin{minipage}[t]{0.325\linewidth}
        \centering
        \includegraphics[width=\columnwidth]{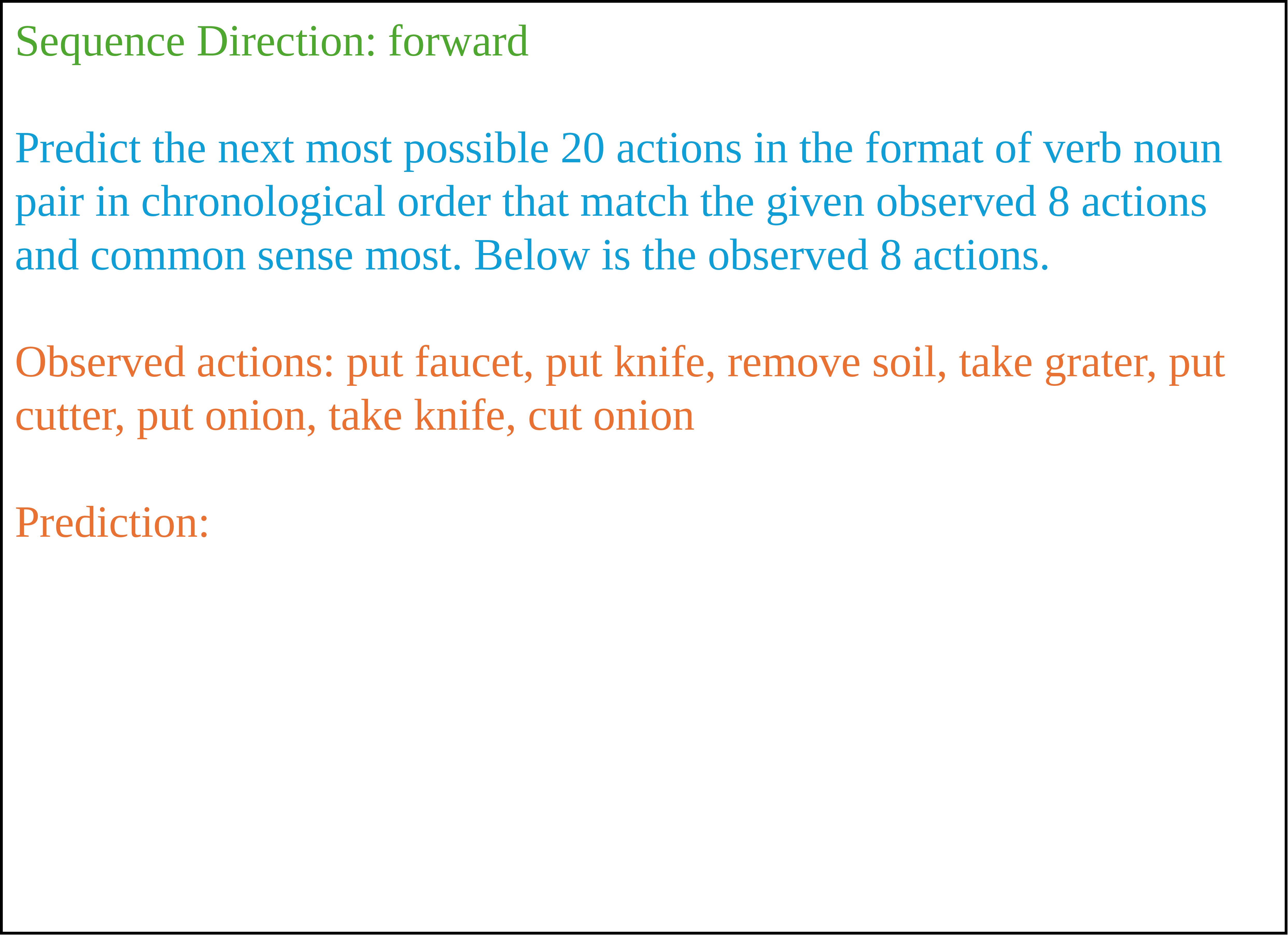}
        \subcaption{Forward action sequence learning}
        \label{fwd_prompt}
      \end{minipage}
  %
      \begin{minipage}[t]{0.325\linewidth}
        \centering
        \includegraphics[width=\columnwidth]{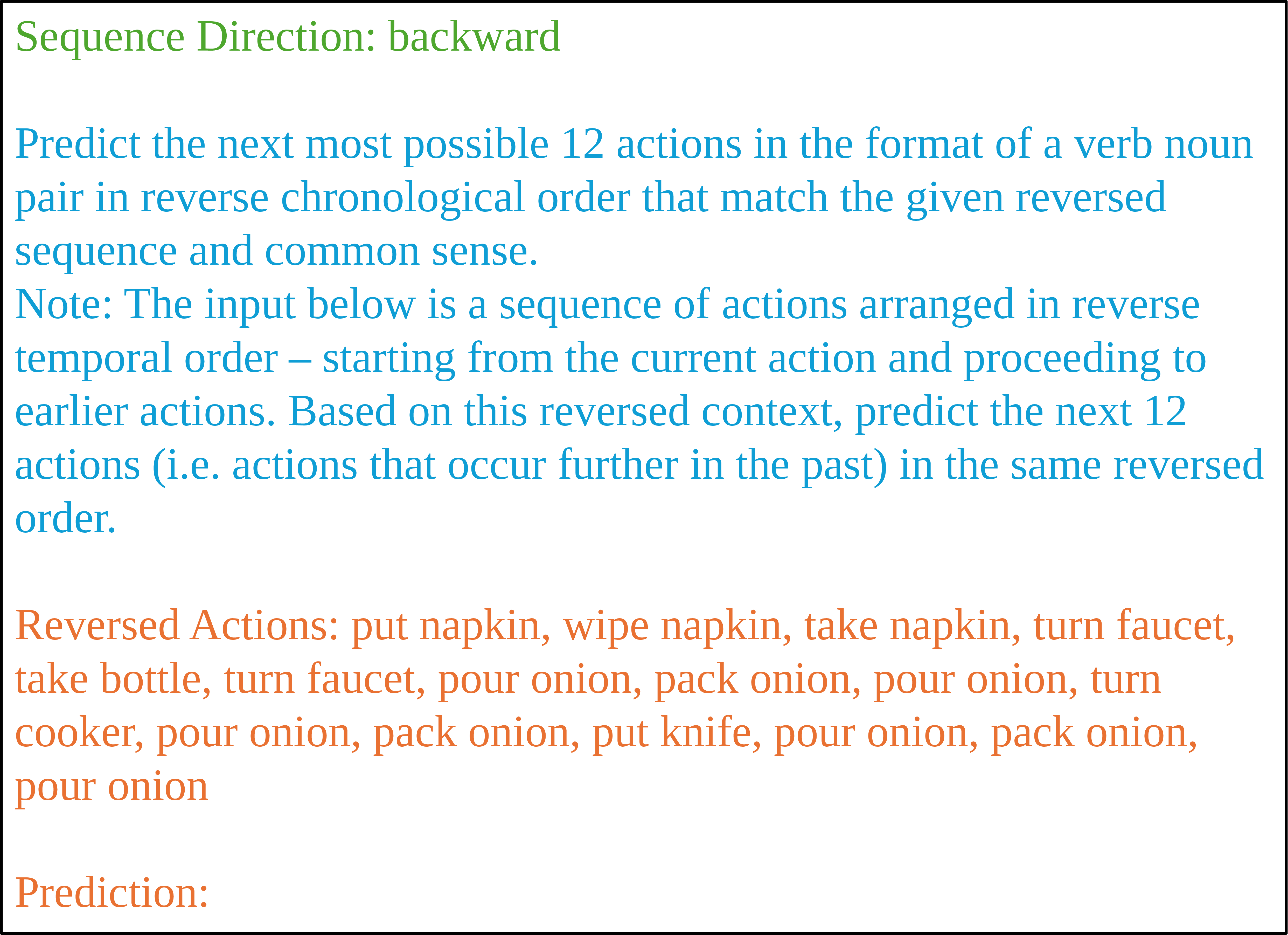}
       \subcaption{Backward action sequence learning}
        \label{bwd_prompt}
      \end{minipage}
  %
      \begin{minipage}[t]{0.325\linewidth}
        \centering
        \includegraphics[width=\columnwidth]{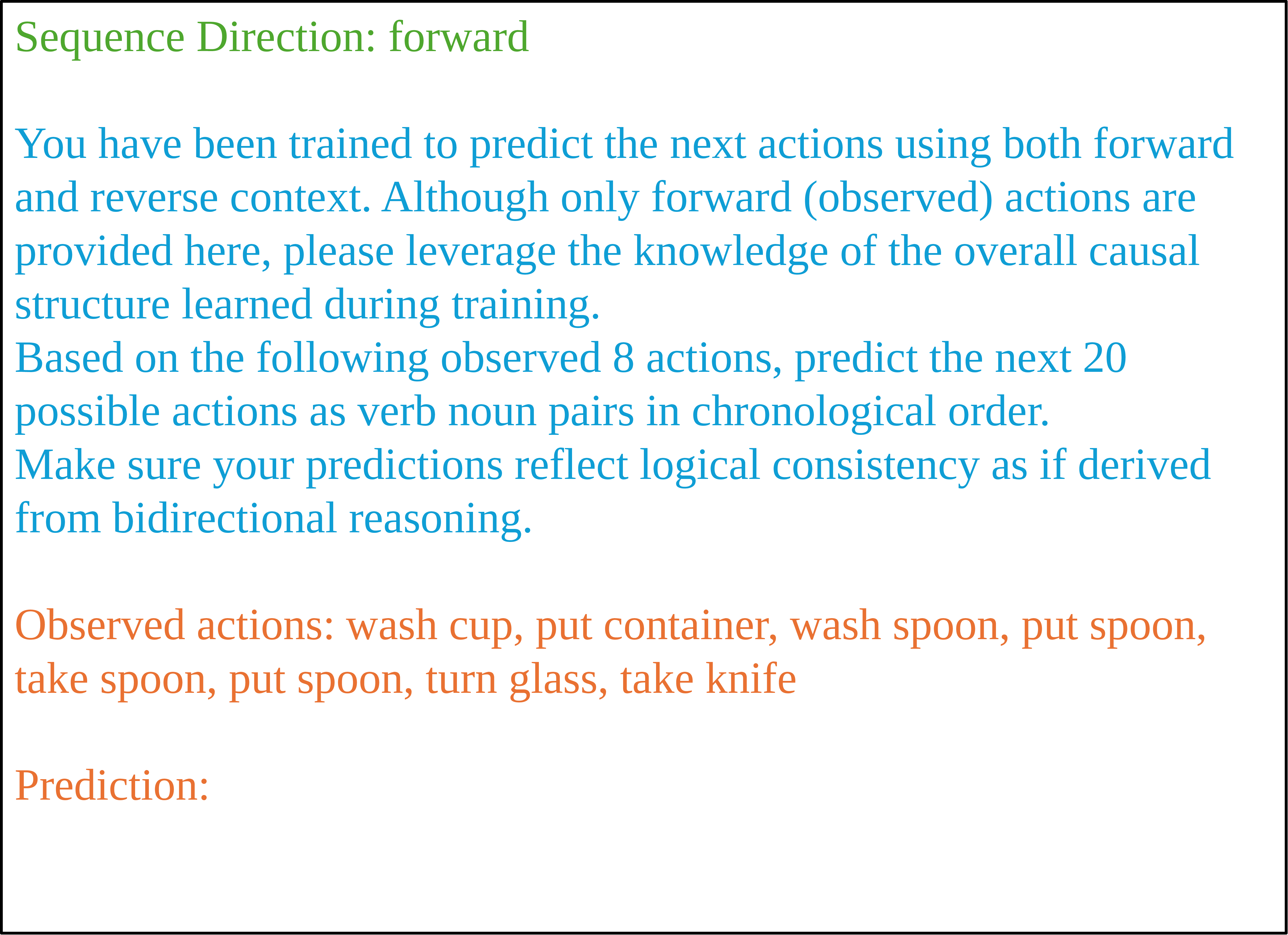}
       \subcaption{For inference}
        \label{inference_prompt}
      \end{minipage}
  
    \end{tabular}
      }
        \vspace{-2mm} 
    \caption{\textbf{
    Prompt with detailed task description.}
    Examples of prompts used for (a) forward and (b) backward action sequence learning during training, and (c) forward-only inference.  
    The prompt contains a detailed description of the task.
    \color{black}
    }
    \label{supp_prompt_design}
  \end{figure*}
\begin{table}[t]
  \renewcommand{\arraystretch}{1.2}
  \caption{
  Ablation study on the task description added to the beginning of the prompt. The best accuracy is highlighted in bold. The optimal accuracy is achieved when the special token is used.
  }
  \begin{center}
  \scalebox{0.8}{
    \begin{tabular}{c | c c c}
      \hline
      \hline
      \makebox[20mm]{Token type} &
      \makebox[10mm]{verb $\downarrow$} & \makebox[10mm]{noun $\downarrow$} & \makebox[10mm]{action $\downarrow$} \\
      \hline
      Detailed description & \bf{0.6781} & 0.6419 & 0.8853 \\
      Special token & 0.6841 & \bf{0.6378} & \bf{0.8741} \\
      \hline
      \hline
    \end{tabular}
    }
    \label{ablation_study_on_token_type}
  \end{center}
\end{table}

\subsection{Task description token at the beginning of the prompt}
The proposed method inserts a special token at the start of the input prompt to clearly indicate the forward anticipation task and the backward anticipation task.
The method employs [forward] for the forward anticipation task and [backward] for the backward anticipation task.
This ablation study compares the accuracy achieved with a detailed description that explains the forward anticipation task and the backward anticipation task.
Figure \ref{supp_prompt_design} presents the input prompt that includes detailed descriptions for the forward anticipation task (refer to Figure \ref{fwd_prompt}) and the backward anticipation task (refer to Figure \ref{bwd_prompt}), and forward-only inference (refer to Figure \ref{inference_prompt}).
The study investigates whether the special token improves task-specific behavior learning in large language models.
Table \ref{ablation_study_on_token_type} presents the evaluation results.
The results indicate that the special token attains the highest accuracy for the action metric.

\subsection{The value of actively utilizing backward anticipation task}
The results indicate that the optimal settings in the paper are an observation interval length of 16 and loss weights of $\alpha = 1.0$ and $\beta = 1.0$, and using the special token is effective. These settings were used in the experiments of the main paper.
These findings indicate that the backward anticipation task does not hinder better learning and that actively utilizing its configuration is effective in conventional models.

%
%
{
    \small
    \bibliographystyle{plain}
    \bibliography{main}
}